\documentclass{article}
\usepackage{ijcai21}

\usepackage{times}
\usepackage[ruled, vlined, linesnumbered]{algorithm2e}
\usepackage{stfloats}
\usepackage{subcaption}
\usepackage{graphicx}
\usepackage{amsmath}
\usepackage{amsfonts}
\usepackage{multirow}
\usepackage{threeparttable}
\usepackage{booktabs}
\usepackage{enumitem}

\title{Cross-Silo Heterogeneous Model Federated Multitask Learning}

\author{ \large{
Xingjian Cao$^{1}$
\and
Zonghang Li$^{1}$
\and
Gang Sun$^{1,3}$
\and
Hongfang Yu$^{1,2}$
\And
Mohsen Guizani$^{4}$
}
\affiliations \large{
$^1$UESTC, China,
$^2$Pengcheng Laboratory, Shenzhen, China\\
$^3$Agile and Intelligent Computing Key Laboratory of Sichuan Province, Chengdu, China\\
$^4$Machine Learning Department, Mohamed Bin Zayed University Of Artificial Intelligence (MBZUAI), Abu Dhabi, UAE~}\\
\emails
\normalsize{x.cao.flc@gmail.com, zhli0117@163.com, gangsun@uestc.edu.cn, yuhf@uestc.edu.cn, mguizani@ieee.org}}

\begin{document}

\maketitle

\begin{abstract}
Federated learning (FL) is a machine learning technique that enables participants to collaboratively train high-quality models without exchanging their private data. Participants utilizing cross-silo federated learning (CS-FL) settings are independent organizations with different task needs, and they are concerned not only with data privacy but also with independently training their unique models due to intellectual property considerations. Most existing FL methods are incapable of satisfying the above scenarios. 
In this study, we present a novel federated learning method CoFED based on unlabeled data pseudolabeling via a process known as cotraining. CoFED is a federated learning method that is compatible with heterogeneous models, tasks, and training processes. The experimental results suggest that the proposed method outperforms competing ones. This is especially true for non-independent and identically distributed settings and heterogeneous models, where the proposed method achieves a 35\% performance improvement.
\end{abstract}

\section{Introduction}
Federated learning (FL) allows different participants to collaborate in solving machine learning problems under the supervision of a center without disclosing participants' private data \cite{kairouz2021advances}. The main purpose of FL is to achieve improved model quality by leveraging the multiparty knowledge from participants' private data without the disclosure of data themselves.

FL was originally proposed for training a machine learning model across a large number of users' mobile devices without logging their private data to a data center \cite{konevcny2015federated}. In this scenario, a center orchestrates edge devices to train a global model that serves a global task. However, some new FL settings have emerged in many fields, including medicine \cite{rieke2020future}, \cite{xiao2021federated}, \cite{raza2022designing}, \cite{courtiol2019deep}, \cite{adnan2022federated}, finance \cite{li2020preserving}, \cite{gu2021privacy}, and network \cite{zhang2021survey}, \cite{nguyen2021federated}, \cite{ghimire2022recent}, \cite{regan2022federated}, where the participants are likely companies or organizations. Generally, the terms \textit{cross-device federated learning} (CD-FL) and \textit{cross-silo federated learning} (CS-FL) can refer to the above two FL settings \cite{kairouz2021advances}. However, the majority of these studies' contributions concern training reward mechanisms \cite{tang2021incentive}, topology designs \cite{marfoq2020throughput}, data protection optimization approaches \cite{zhang2020batchcrypt}, etc., and for their core CS-FL algorithms, they simply follow the idea of gradient aggregation used in CD-FL. Such studies ignore the fact that organizations or companies, as participants, may be more heterogeneous than device participants.

One of the most important heterogeneities to address under the CS-FL setting is model heterogeneity; that is, models may have different architecture designs. Different devices in CD-FL typically share the same model architecture, which is given by a center. As a result, the models obtained through local data training on different devices differ only in terms of their model parameters, allowing the center to directly aggregate the gradients or model parameters uploaded by the participating devices. However, participants in a CS-FL scenario are usually independent companies or organizations, and they are capable of designing unique models. They prefer to use their own independently designed model architectures rather than sharing the same model architecture with others. At this time, the strategy used in CD-FL cannot be applied to CS-FL models with different architectures. Furthermore, model architectures may also be intellectual properties that need to be protected, and companies or organizations that own these properties do not want them to be exposed to anyone else, which makes model architecture sharing hard for CS-FL participants to accept. An ideal CS-FL method should treat each participant's model as a black box, without the need for its parameters or architecture.

The heterogeneity among models results in the need for training heterogeneity. Under the CD-FL setting, participants usually train their models according to the configuration of a center, which may include training algorithms (such as stochastic gradient descent) and parameter settings (such as the learning rate and minibatch size). However, when participants like companies or organizations use their independently designed model architectures in CS-FL, they need to choose different local training algorithms that are suitable for their models and exert full control over their local training processes. Decoupling the local training processes of different participants not only enables them to choose suitable algorithms for their models but also prevents the leakage of their training strategies, which may be their intellectual properties.

In addition, CS-FL is more likely to face task heterogeneity than CD-FL. In CD-FL scenarios, all devices usually share the same target task. In terms of classification tasks, the task output categories of all devices are exactly the same. Under the CS-FL setting, because the participating companies or organizations are independent of each other and have different business needs, their tasks may be different. Of course, we must assume that there are still similarities between these tasks. In terms of classification tasks, the task output categories of different participants in CS-FL may be different. For example, an autonomous driving technology company and a home video surveillance system company both need to complete their individual machine learning classification tasks. Although both of them need to recognize pedestrians, the former must also recognize vehicles, whereas the latter must recognize indoor fires. Therefore, the task output categories of the autonomous driving technology company include pedestrians and vehicles without indoor fires, whereas the task output categories of the home video surveillance system company include pedestrians and indoor fires without vehicles. It is easy to see that, in contrast to the complete task consistency of CD-FL, CS-FL participation by independent companies or organizations is more likely to encounter situations in which the different participants possess heterogeneous tasks.

\begin{figure}[t]
  \centering
  \begin{subfigure}{\columnwidth}
    \centering
    \includegraphics[width=\linewidth]{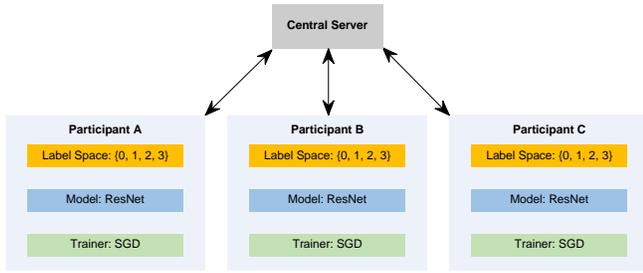}
    \caption{CS-FL setting without heterogeneity}
  \end{subfigure}%
  \par\bigskip
  \begin{subfigure}{\columnwidth}
    \centering
    \includegraphics[width=\linewidth]{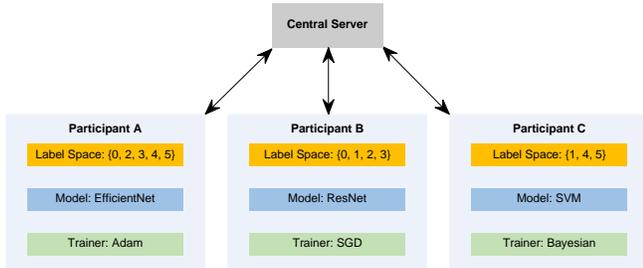}
    \caption{CS-FL setting with heterogeneity (HFMTL)}
  \end{subfigure}%
  \caption{(b) is an example of an HFMTL setting with 3 participants. A central server needs to coordinate 3 participants to solve their classification tasks via federated training. Compared with (a), which has no heterogeneity, (b) also contains different machine learning model architectures and training optimization algorithms, and the tasks of participants are distinct from one another(i.e., different label spaces).}
  \label{fig1}
\end{figure}

Although the number of participants in CS-FL is much smaller than that in CD-FL in general, the heterogeneity of the participants, including model heterogeneity, training heterogeneity and task heterogeneity, may bring more challenges. Overall, we use the term \textit{heterogeneous federated multitask learning} (HFMTL) to refer to the FL settings that contain the above three heterogeneity requirements, and Fig. \ref{fig1}(b) shows an example of the HFMTL setting with three participants. Different participants may have different label spaces. For example, Participant $B$ has a label space $\left\{0, 1, 2, 3\right\}$. This space can be different from those of other participants, e.g., Participant $C$ has $\left\{1, 4, 5\right\}$. The classification task of Participant $B$ is to classify inputs with the labels in $\left\{0, 1, 2, 3\right\}$, while Participant $C$ aims to classify inputs with the labels in $\left\{1, 4, 5\right\}$. Therefore, they have different tasks and usually have different categories of local training data.

Our main motivation in this paper is to address the needs of \textit{model heterogeneity}, \textit{training heterogeneity}, and \textit{task heterogeneity} in CS-FL settings. Aiming at the HFMTL scenario with these heterogeneities, we propose a novel FL method. The main contributions of this paper are as follows.

\begin{itemize}
\item We propose an FL method that is simultaneously compatible with heterogeneous tasks, models, and training processes, and it boost the performance of each participant's model.
\item Compared with the existing FL methods, the proposed method not only protects private local data but also protects the architectures of private models and private local training methods.
\item Compared with these other methods, the proposed method achieves more performance improvements for the non-independent and identically distributed (non-IID) data settings in FL and can be completed in one round.
\item We conduct comprehensive experiments to corroborate the theoretical analysis conclusions and the impacts of different settings on the suggested method.
\end{itemize}

The rest of this paper is structured as follows. Section 2 provides a brief overview of relevant works. Section 3 contains the preliminaries of our method. Section 4 describes the proposed method in detail. Section 5 offers the experimental results and analysis. Section 6 summarizes the paper.

\section{Preliminaries}
We first formulate the HFMTL problem and introduce the cotraining method that inspires us to propose the communication-efficient FL (CoFED) method.
\subsection{HFMTL}
An FL setting contains $N$ participants, and each of them has its own classification task $T_i$, input space $\mathcal{X}_i$, output space $\mathcal{Y}_i$, and $\mathcal{D}_i$ consisting of all valid pairs $(\boldsymbol{x}, y)$, where $\boldsymbol{x} \in \mathcal{X}_i, y \in \mathcal{Y}_i$. Since we indicate that $T_i$ is a classification task, $\mathcal{Y}_i$ is the label space of $T_i$. Each participant trains its machine learning model with supervised learning to perform a classification task, so each participant has its own local data $D_i \subset \mathcal{D}_i$.

Since the HFMTL settings contain heterogeneous tasks and heterogeneous models $f$, we can assume that for the general $i \ne j$:
\begin{equation}
\mathcal{Y}_i\ne \mathcal{Y}_j, f_i\ne f_j
\end{equation}
On the other hand, the classification tasks of each participant should have commonality with the tasks of other participants. In our setting, this commonality manifests as the overlap between the label spaces. This means that:
\begin{equation}
    \forall \mathcal{Y}_i\rightarrow \left\{ j|\mathcal{Y}_i\cap \mathcal{Y}_j\ne \oslash , i\ne j \right\} \ne \oslash 
\end{equation}

Assuming that a model $f_i$ has been trained for $T_i$, its generalization accuracy can be defined as:
\begin{equation}
    GA(f_i) = \mathbf{E}_{(\boldsymbol{x}, y) \sim \mathcal{D}_i}[\mathbf{I}\left( f_i\left( \boldsymbol{x} \right) \equiv y \right)]
\end{equation}
where $\mathbf{E}[\cdot]$ is the expectation and $\mathbf{I}(\cdot)$ is an indicator function: $\mathbf{I}(\text{TRUE}) \equiv 1 $ and $\mathbf{I}(\text{FALSE}) \equiv 0 $.

The goal of this paper is to propose an HFMTL method that is compatible with heterogeneous tasks, models, and training processes. This method should help to improve the performance of each participant model without sharing the local private datasets $D_i$ and private models $f_i$ of these participants:
\begin{equation}
    f_i^{fed} = \underset{f_i}{arg\max}\,\,GA\left( f_i \right) 
\end{equation}
In addition, assuming that the model locally trained by the participant is $f_{i}^{loc}$, we expect that the model $f_{i}^{fed}$ trained with our FL method yields better performance for each participant:
\begin{equation}
    GA(f_{i}^{fed}) > GA(f_{i}^{loc}), i=1,2,\cdots,N
\end{equation}

\subsection{Cotraining}
In HFMTL settings, many tricky problems come from the differences between participant models and tasks. However, cotraining is an effective method for training different models. Therefore, ideas that are similar to cotraining can be used to solve problems under HFMTL settings.

Cotraining is a semi-supervised learning technique proposed by Blum \textit{et al.} \cite{blum1998combining} that utilizes unlabeled data from different views. The \textit{view} means a subset of instance attributes, and two classifiers that can be obtained by training on different view data. Cotraining lets the two classifiers jointly give some pseudo-labels for unlabeled instances, and then retraining the two classifiers on the original training set and the pseudo-labeled dataset can improve the classifier performance. Wang \textit{et al.} \cite{wang2007analyzing} pointed out that cotraining is still effective for a single view, and applying cotraining can bring greater classifier performance improvement when  the divergence of different classifiers is larger. In FL, local models of different participants are trained from their local data. Under the HFMTL setting, due to differences in model architecture and data distribution, models of different parties are likely to have large divergences, which is beneficial for applying Cotraining.

\section{Related Work}
The federated average (FedAvg) algorithm was originally proposed by McMahan \textit{et al.} \cite{mcmahan2017communication} to solve machine learning federated optimization problems on mobile devices. The core idea of FedAvg method is to pass model parameters instead of private data from different data sources, and use a weighted average of model parameters from different data sources as a global model. In each round of communication in the FedAvg method, the central server broadcasts the global model to participants, and then the participants who received the global model continue to train the global model using local private data. Following the local training phase, participants submit trained models to the center which utilizes the weighted average of different participant models as the global model for the next round of communication.

Inspired by the FedAvg algorithm, many FL methods \cite{wang2019adaptive}, \cite{li2020federatedhn} based on the aggregation of model parameters have been proposed. These methods are mainly suitable for federated training under the CD-FL setting where the tasks and model architectures are usually published by a central server, and all participants (i.e. devices) sharing the same model architecture makes aggregation of model parameters an effective knowledge sharing strategy. In addition, under the CD-FL setting, it is also practical for the central server to control the local training process and parameters (such as learning rate, epoch number, and mini-batch size, etc.). However, under the CS-FL setting where the model architectures and tasks of different participants may be different, and the training process and parameters are reluctant to be controlled by the center, these FL methods based on aggregation of model parameters are generally unable to cope with these heterogeneity challenges. FedProx proposed by Li \textit{et al.} \cite{li2020federatedhn} introduced a proximal term to overcome statistical heterogeneity and systematic heterogeneity (i.e., stragglers), but FedProx is unable to cope with heterogeneous model architectures due to model parameter aggregation.

In recent years, personalized federated learning has been proposed to address the personalized needs of participants. Smith \textit{et al.} \cite{smith2017federated} proposes a multitask FL framework MOCHA which clusters different tasks according to their relevance by an estimated matrix. Khodak \textit{et al.} \cite{khodak2019adaptive} proposed an adaptive meta-learning framework utilizing online learning ARUBA, which has the advantage of eliminating the need for hyperparameter optimization during personalized FL. Lin \textit{et al.} \cite{lin2020meta} and Fallah \textit{et al.} \cite{fallah2020personalized} proposed a model-agnostic meta-learning method and their variants to achieve personalized federated learning. Dinh \textit{et al.} \cite{dinh2020personalized} proposed a personalized FL method based on meta-learning by Moreau envelope, which leverages the $l2$-norm regularization loss to balance the personalization performance and generalization performance.

It should be pointed out that personalized tasks are different from heterogeneous tasks. Tasks with personalized settings always have the same label spaces, while tasks with heterogeneous settings have different label spaces. A typical example of personalized tasks that can be presented from \cite{smith2017federated} is to classify users' activities by using their mobile phone accelerometer and gyroscope data. For each user, the model needs to provide outputs from the same range of activity categories, but the classification for each user is regarded as a personalized task due to the differences in their heights, weights, and personal habits. Despite these differences, from the data distribution perspective, the data of heterogeneous classification tasks can be regarded as non-IID sampling results from an input space $\mathcal{X}$, which contains all instances of all labels in a label space $\mathcal{Y}$, and $\mathcal{Y}$ is the union of the label spaces of all participants. Each participant only has instances of some categories in $\mathcal{Y}$ since its label space is a subset of $\mathcal{Y}$. Therefore, an FL method that is compatible with personalized tasks can also be used for heterogeneous tasks.

However, the core idea of all these methods is model parameter aggregation, which requires the model architectures of all participants to be consistent or only partially different. However, under CS-FL settings, participants as companies or organizations usually need to use their own independently designed model architectures. Li \textit{et al.} \cite{li2019fedmd} proposed a FL method FedMD leveraging model distillation to transfer the knowledge of different participant's model by aligning the output of neural networks models on a public dataset. Although FedMD participants are allowed to use neural network models of different architectures, they must be consistent in the logit output layer, and FedMD is not compatible with participants using non-neural network models. Also, FedMD needs a large number of labeled data for participant model alignment, which raises the bar for its application.

Model parameter aggregation also leads to the leakage of participant models. Many studies try to overcome model leakage using various secure computing methods including differential privacy \cite{wei2020federated}, \cite{hu2020personalized}, \cite{adnan2022federated}, secure multiparty computing \cite{yin2020fdc}, \cite{liu2020secure}, homomorphic encryption \cite{jia2021blockchain}, \cite{fang2021privacy}, \cite{zhang2020batchcrypt}, blockchain \cite{li2022blockchain}, \cite{otoum2022federated}, and trusted execution environments \cite{mo2021ppfl}, \cite{chen2020training}, However, these technologies still have certain drawbacks, such as high computational costs or hardware specific.

\section{Methodology}
\subsection{The Overall Steps of CoFED}
The main incentive of FL is that it improves the performance of participant models. The poor performance of locally trained models is mainly due to insufficient local training data, so these models fail to learn enough task expertise. Hence, to increase the participant model performance, it is vital to allow them to learn from other participants. The most popular and straightforward methods of sharing knowledge are shared models or data, however both are forbidden by FL settings, thus we must devise alternative methods.

A research \cite{wang2007analyzing} discovered that a sufficiently enough amount of variety across the models is necessary to increase classification model performance. Generally, it is difficult to generate models with large divergences under single-view settings. However, under FL settings, The architectures of the participant models may be highly varied, and they may have been trained on distinct local datasets that are very likely to be distributed differently. All of these may increase the diversity of different participants' models. Therefore, if enough unlabeled data are obtained, we can regard the federated classification problem as a semisupervised learning problem, and it is very suitable to adopt cotraining-like techniques due to the high diversity between different participant models. We provide the overall steps of CoFED as follows.

\begin{enumerate}
    \item \textbf{Local training}: Each participant independently trains a local model on its private dataset.
    \item \textbf{Pseudolabeling}: Each participant pseudolabels an unlabeled dataset, which is public to all participants, with its locally trained model.
    \item \textbf{Pseudolabel aggregation}: Each participant uploads its pseudolabeling results to a central server, and the center votes for the pseudolabeled dataset with high confidence for each category based on the category overlap statuses of different participants and the pseudolabeling results. After that, the center sends all pseudolabels to each participant.
    \item \textbf{Update training}: Each participant trains its local model on the new dataset created by combining the local dataset with the received pseudolabeled dataset.
\end{enumerate}

It can be seen that there are some differences between CoFED and the cotraining process under single-view settings. First, in cotraining, the training sets used by different classifiers are the same, while the training sets of the different classifiers in CoFED come from different participants, so they are usually different. Second, the target tasks of different classifiers in cotraining are the same; that is, they have the same label space. However, in CoFED, the label spaces of different classifiers are different, and the pseudolabeling of unlabeled samples need to be performed according to the pseudolabel aggregation results of the overlapping classification process. Furthermore, cotraining completes training by repeating the above process, while the process is performed only once in CoFED.

\subsection{Analysis}

Suppose that we are given two binary classification models $f$ and $g$ from a hypothesis space $\mathcal{H}: \mathcal{X} \rightarrow \mathcal{Y}, |\mathcal{H}| < \infty$, and an oracle model $h \in \mathcal{H}$ whose generalization error is zero. We can define the generalization disagreement between $h_1 \in \mathcal{H}$ and $h_2 \in \mathcal{H}$ as:
\begin{equation}
    \begin{aligned}
     d(h_1,h_2)&=d(h_1,h_2|\mathcal{X}) \\
               &=\mathbf{Pr}(h_1(x) \ne h_2(x) | x \in \mathcal{X})
    \end{aligned}
\end{equation}

Therefore, the generalization errors of $f$ and $g$ can be computed as $d(f,h)$ and $d(g,h)$, respectively. Let $\varepsilon$ bound the generalization error of a model, and let $\delta>0$; a learning process generates an approximate model $h'$ for $h$ with respect to $\varepsilon$ and $\delta$ if and only if:
\begin{equation}
    \mathbf{Pr}(d(h', h) \ge \varepsilon) \le \delta
\end{equation}

Since we usually have only a training dataset $X \subset \mathcal{X}$ containing finite samples, the training process minimizes the disagreement over $X$:
\begin{equation}
    \mathbf{Pr}(d(f,h|X) \ge \varepsilon) \le \delta
\end{equation}

\textbf{Theorem 1.} \textit{
Given that $f$ is a probably approximately correct (PAC) learnable model trained on $L \subset \mathcal{D}$, $g$ is a PAC model trained on $L_{g} \subset \mathcal{D}$, and $\varepsilon_f < \frac{1}{2}$, $\varepsilon_g < \frac{1}{2}$. $f$ and $g$ satisfy that the following:
\begin{align}
    \label{9}
    & \mathbf{Pr}(d(f, h) \ge \varepsilon_f) \le \delta \\
    \label{10}
    & \mathbf{Pr}(d(g, h) \ge \varepsilon_g) \le \delta
\end{align}
If we use $g$ to pseudolabel an unlabeled dataset $X_u \subset \mathcal{X}$, we generate a pseudolabeled dataset
\begin{equation}
    P=\left \{ (x,y)|x \in X_u, y=g(x) \right \}
\end{equation}
and combine $P$ and $L$ into a new training dataset $C$. After that, $f'$ is trained on $C$ by minimizing the empirical risk. Moreover,}
\begin{align}
    \label{13}
    & |L| \varepsilon_f < \sqrt[|P|\varepsilon_g]{(|P|\varepsilon_g)!} \hspace{1mm} e-|P|\varepsilon_g \\
    \label{14}
    & \varepsilon_{f'} = \max \left \{\varepsilon_f + \frac{|P|}{|L|}(\varepsilon_g-d(g,f')), 0 \right \}
\end{align}
\textit{where $e$ is the base for natural logarithms; then,}
\begin{equation}
    \label{15}
    \mathbf{Pr}(d(f',h) \ge \varepsilon_{f'}) \le \delta
\end{equation}

Theorem 1 has been proven in \cite{wang2007analyzing}. Assume that $f$ and $g$ are 2 models from different participants that satisfy (\ref{9}) and (\ref{10}). The right side of (\ref{13}) monotonically increases as $|P|\varepsilon_g \in (0, \infty)$, which indicates that a larger pseudolabeled dataset $P$ enables a larger upper bound of $|L| \varepsilon_f$. That is, if $f$ is a model trained on a larger training dataset with higher generalization accuracy, a larger unlabeled dataset is required to further improve the generalization accuracy of $f$.It can be seen from (\ref{14}) and (\ref{15}) that when $d(g,f')$ is larger, the lower bound of the generalization error of $f'$ under the same confidence is smaller, which is because that $f'$ is trained on $L$ and $P$, and $P$ is generated by $g$, $d(g,f')$ mainly depends on  the degree of divergence between $f$ and $g$, i.e., the difference between the training dataset of $f$ and $g$. In FL settings, substantial diversity across different local training datasets is fairly prevalent, hence the performance improvement requirement is generally satisfied. The same conclusion applies to the boost version $g'$ obtained by switching $f$ and $g$ due to symmetry.

On the other hand, if $|P|$ is sufficiently large since $P$ is generated by $g$, the $f'$ trained on $C$ can be treated as proximal to $g$. That is, if we repeat the above process on $f'$ and $g$, $f'$ and $g$ may be too similar to improve them. Therefore, we utilize a large unlabeled dataset and only perform the above process once instead of iterating for multiple rounds; this technique can also avoid the computational cost increase caused by multiple training iterations.

An intuitive explanation for the CoFED method is that when the different participant models have great diversity, the differences in the knowledge possessed by the models are greater. As a result, the knowledge acquired by each participant model from others contains more knowledge that is unknown to itself. Therefore, more distinctive knowledge leads to greater performance gains. In addition, when mutual learning between different models is sufficient, the knowledge difference between them will almost disappear, and it is difficult for mutual learning to provide any participant's model with distinctive knowledge.

\subsection{Pseudolabel Aggregation}

After each participant uploads its pseudolabeling results to the public unlabeled dataset, the center exploits their outputs to pseudolabel the unlabeled dataset. In this subsection, we explain the implementation details of this step.

Assume that the union of the label spaces of all participants' tasks is
\begin{equation}
\label{union}
\mathcal{Y}=\bigcup_{i=1}^N{\mathcal{Y}_i=\left\{ c_k \right\} , k=1,2,\cdots,n_c}
\end{equation}
where $n_c$ is the number of elements in the whole label space $\mathcal{Y}$, and each category $c_k$ exists in the label space of one or multiple participants' label spaces:
\begin{equation}
    c_k\in \bigcap_{j=1}^{m_k}{\mathcal{Y}_{i_j}, 1\leq i_1 < i_2 < \cdots< i_{m_k}\leq N}
\end{equation}
$m_k$ is the number of participants who possess category $c_k$. At the same time, we define the pseudolabeled dataset $P_k$ for category $c_k$, and $P_k$ is used to store the indices of the instances in the public dataset corresponding to each category $c_k$ after pseudolabel aggregation.

For each category $c_k$ existing in $\mathcal{Y}_{i_j}$, the model $f_{i_j}$ classifies the instances in the public unlabeled dataset $D^{pub}$ as belonging to category $c_k$, and the set of these instances can be defined as
\begin{equation}
    S_j=\left\{ x|f_{i_j}\left( x \right) \equiv c_k,x\in D^{pub} \right\} , j=1,2,\cdots,m_k
\end{equation}

For an instance $x\in D^{pub}$, if we regard the outputs of different participant models on $x$  for category $c_k$ as the outputs of a two-class (belonging to category $c_k$ or not) ensemble classifier $g$, since (\ref{14}) suggests that a lower generalization error bound $\varepsilon_g$ is helpful, we can set a hyperparameter $\alpha$ to make the results more reliable. That is, if
\begin{equation}
    \frac{|\left\{ S_j|x\in S_j \right\} |}{m_k} > \alpha, 0 \le \alpha \le 1,
\end{equation}
an $\alpha$ value of 0 means that whether $x$ is marked as belonging to category $c_k$ requires only one participant to agree, while an $\alpha$ of 1 means that the consent of all participants is required. After that, we can put the index of $x$ into $P_k$. After all $P_k$ are generated, the central server sends the corresponding pseudolabeled dataset to each participant's task based on its label space $\mathcal{Y}_i$. For the participants whose label space is $\mathcal{Y}_i$, the corresponding pseudolabeled dataset received from the center is:
\begin{equation}
    R_i = \{(P_k, c_k)|c_k \in \mathcal{Y}_i\}
\end{equation}
Assuming that $C_i[index]$ stores the results of $f_i(D^{pub}[index])$, $\mathcal{P}=\{P_k|c_k \in \mathcal{Y}\}$ and $M=|D^{pub}|$, Algorithm 1 describes the above process.
\begin{algorithm}[!t]
\DontPrintSemicolon
 \caption{Pseudolabel aggregation}
 \KwIn{$\{C_i\}$, $\mathcal{Y}_i$, $\mathcal{Y}$, $\alpha$, $M$}
 \KwOut{$\{P_k\}$}
 \SetKwBlock{Begin}{function}{end function}
 \Begin(\text{Aggregation} {$(\{C_i\}$, $\mathcal{Y}_i$, $\mathcal{Y}$, $\alpha$, $M)$})
 {
  \tcp*[l]{Initialization}
  $TOTAL$ = empty dict\;
  $\mathcal{P}$ = empty set\;
  
  \tcp*[l]{Counting $c_k$}
  \ForAll {$c_k \in \mathcal{Y}$}
  {
    $TOTAL[c_k] = |\{i|c_k \in \mathcal{Y}_i\}|$\;
    $\mathcal{P}[c_k]$ = empty set\;
  }
  
  \tcp*[l]{Label aggregating}
  \ForAll {$index = 1$ \textbf{to} $M$}
  {
    $COUNT$ = empty dict with default value 0\;
    \ForAll {$i = 1$ \textbf{to} $N$}
    {
        $COUNT[C_i[index]] = COUNT[C_i[index]] + 1$\;
    }
    \ForAll {$c_k \in keys(COUNT)$}
    {
        \uIf {$\frac{COUNT(c_k)}{TOTAL(c_k)}>\alpha$}
        {
            add $index$ to $P_k$\;
        }
        \Else
        {
           continue\;
        }
        
    }
  }

 \Return{$\{P_k\}$}
 }
 \end{algorithm}

It should be pointed out that some indices of $x \in D^{pub}$ may exist in multiple $P_k$ because all $\mathcal{Y}_i$ are different from each other. Therefore, the different $P_k$ in $R_i$ may overlap, resulting in contradictory labels. To build a compatible pseudolabeled dataset $R_i$, the indices contained by different $P_k$ should be removed from $R_i$.

\subsection{Unlabeled Dataset}
To perform the CoFED method, we need to build a public unlabeled dataset for all participants. Although an unlabeled dataset that is highly relevant to the classification tasks is preferred, we find that even less relevant datasets can yield sufficient results in our experiments. For the image classification tasks that we focus on, the almost unlimited image resources that are publicly accessible on the Internet can be built into an unlabeled dataset. Another benefit of utilizing resources that each participant can independently obtain is that this strategy can prevent the distribution of unlabeled datasets by a central server, thereby saving the limited communication resources in an FL scenario.

The reason why less relevant datasets work is that even though different objects may have commonality, we can use this commonality as the target of the tasks involving different objects. For example, if someone asks you what an apple looks like when you have nothing else but a pear, you might tell the person that it looks similar to a pear. This may give him or her some incorrect perceptions about apples, but it is better than nothing, and this person will at least be less likely to recognize a banana as an apple. That is, although you have not been able to tell him or her exactly what an apple looks like, the process still improves his or her ability to recognize apples. For a similar reason, even if a less relevant unlabeled dataset is used to transfer knowledge, it can also yield improved model performance in FL scenarios.

\subsection{Training Process}
In the CoFED method, each participant needs to train its model twice, i.e., local training and update training. Both training processes are performed locally, where no exchange of any data with other participants or the central server is necessary. The benefits of this approach are as follows.
\begin{enumerate}
    \item It prevents the leakage of participant data, including the local data and models.
    \item It avoids the loss of stability and performance through communication.
    \item It decouples the training processes of different participants so that they can independently choose the training algorithms and training configurations that are most suitable for their models.
\end{enumerate}

\subsection{Different Participant Credibility Levels}
In practical applications, the problem of different participant credibility levels may be encountered, resulting in uneven pseudolabel quality for different participants. Credibility weights can be assigned to the pseudolabels provided by different participants. Accordingly, Algorithm 1 can be modified to calculate $TOTAL(c_k)$ and $COUNT(c_k)$ by adding the weights of different participant, and the unmodified version of Algorithm 1 is equivalent to the case in which the weight of each participant is 1.

Different bases can be used for setting the weights. For example, since the quality of the model trained on a larger training set is generally higher, weights based on the size of the utilized local dataset may be helpful for the unbalanced data problem. The test accuracy can also be used as a basis for the weights, but the participants may be reluctant to provide this information. Therefore, we can make decisions according to the actual scenario.

\subsection{Non-IID Data Settings for Heterogeneous Tasks}
Data can be non-IID in different ways. We have pointed out that the heterogeneous task setting itself is an extreme non-IID case of the personalized task setting. However, under the heterogeneous task setting, the instance distributions of a single category in the local datasets of different participants can still be IID or non-IID. The IID case means that the instances of this category owned by different participants have the same probability distribution, while in the extreme non-IID case, each participant may only have the instances of one subclass of this category, and different participants have different subclasses. A non-IID example is a case in which pets are contained in the label spaces of two participants, but all pet instances in the local training set of one participant are dogs, while the other participant only has cat images in its local dataset.

The existing FL methods based on parameter aggregation usually work well for IID data but suffer when encountering non-IID data. Zhao \textit{et al.} \cite{zhao2018federated} showed that the accuracy of convolutional neural networks (CNNs) trained with the FedAvg algorithm could be significantly reduced, by up to 55\%, with highly skewed non-IID data. Since non-IID data cannot be prevented in practice, addressing them has always been regarded as an open challenge in FL \cite{kairouz2021advances}, \cite{li2020federated}. Fortunately, in the CoFED method, where model diversity is helpful for improving performance, a non-IID data setting is usually beneficial. This is because models trained on non-IID data generally have more divergences than models trained on IID data.

\section{Experiments}

In this section, we execute the CoFED method under different FL settings to explore the impacts of different conditions and compare it with existing FL methods. The source code can be found at https://github.com/flcao/CoFED.

\subsection{Model Settings}
The CoFED method enables participants to independently design different models. For example, some participants use CNNs as classifiers, while others use support vector machine (SVM) classifiers. We use randomly generated 2-layer or 3-layer CNNs with different architectures as the models for the different participants in image classification tasks to illustrate that CoFED is compatible with heterogeneous models. Ten of the 100 employed model architectures are shown in Table \ref{netStruct}.

\begin{table}[ht]
\caption{Network Architectures}
\label{netStruct}
\centering
\begin{tabular}{cccc}
\toprule
Model & \begin{tabular}[c]{@{}c@{}}1st \\ conv layer\end{tabular} & \ \begin{tabular}[c]{@{}c@{}}2nd \\ conv layer\end{tabular} & \ \begin{tabular}[c]{@{}c@{}}3rd \\ conv layer\end{tabular} \\ \midrule
1 & 24 3x3 & 40 3x3 & none \\
2 & 24 3x3 & 32 3x3 & 56 3x3 \\ 
3 & 20 3x3 & 32 3x3 & none \\ 
4 & 24 3x3 & 40 3x3 & 56 3x3 \\ 
5 & 20 3x3 & 32 3x3 & 80 3x3 \\ 
6 & 24 3x3 & 32 3x3 & 80 3x3 \\ 
7 & 32 3x3 & 32 3x3 & none \\ 
8 & 40 3x3 & 56 3x3 & none \\ 
9 & 32 3x3 & 48 3x3 & none \\ 
10 & 48 3x3 & 56 3x3 & 96 3x3 \\ \bottomrule
\end{tabular}
\end{table}

To demonstrate that CoFED is applicable to broader ranges of models and task types than other approaches, we choose four types of classifiers as the participant models in the Adult dataset experiment: decision trees, SVMs, generalized additive models, and shallow neural networks.

\subsection{CIFAR Dataset Experiment}
We set the number of participants to 100 in our CIFAR experiments. For each participant, we randomly select a few of the 20 superclasses of CIFAR100 \cite{krizhevsky2009learning} as its label space. Since we try to study the effect of the proposed method on heterogeneous tasks, the label spaces of different participants are generally different, but some overlap may occur. The local dataset of each participant consists of the samples in its label space.

The distributions of each superclass possessed by the different participants who own this superclass sample may encounter two situations. In the first case, we assume that the samples of each participant with the superclass are uniformly and randomly sampled from all samples of the superclass in CIFAR100 (that is, the IID setting). In this case, each participant usually has samples belonging to all 5 subclasses of each superclass in its label space. In the second case, we assume that each participant who owns the superclass only has samples belonging to some of the subclasses of its superclass (in our experiment, 1 or 2 subclasses), that is, the non-IID setting. The details of the experimental data settings are as follows.

\textbf{IID Data Setting}: Each participant is randomly assigned 6 to 8 superclasses in CIFAR100 as its label space. For the local training sets, each participant has 50 instances from each superclass in the CIFAR100 training set, and these 50 samples are evenly sampled from the samples of this superclass in the training set of CIFAR100. No overlap occurs between the training sets of any participants. For the test sets, all instances in the test set of CIFAR100 are used, whereby each participant's test set has 500 instances for each superclass.

\textbf{Non-IID Data Setting}: This setting is almost the same as the IID setting; the difference is that the sample of a superclass of each participant is only randomly acquired from 1 to 2 subclasses of the superclass in the CIFAR100 training set. The configuration of the test set is exactly the same as that used with the IID setting. The non-IID data setting is often regarded as more difficult than the IID one since a model that learns only 1 or 2 subclasses of a superclass during the training process is required to recognize all 5 subclasses included in the test set.

In this experiment, the public unlabeled dataset used in CoFED is the training set of CIFAR10 \cite{krizhevsky2009learning}. Since CoFED is compatible with heterogeneous training processes, we conduct a grid search to determine the optimal training parameter settings for each participant task. The training configuration optimized for each participant (including the trainer, learning rate, and learning rate decay) is used in the initial local training phase, and the update training settings in the final step are adjusted based on these parameters. We always use a minibatch size of 50 for the local training process and 1000 for the update training process.

\begin{figure}[t]
  \centering
  \begin{subfigure}{.5\columnwidth}
    \centering
    \includegraphics[width=\linewidth]{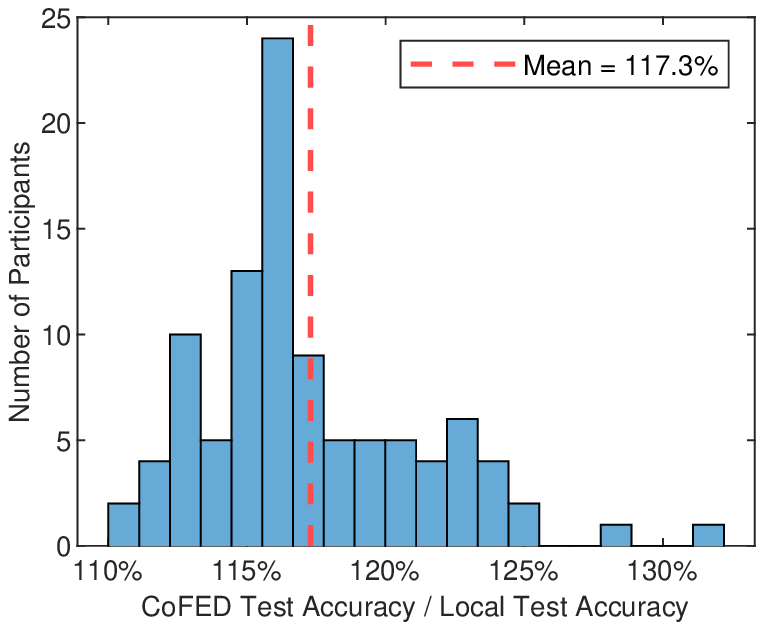}
    \caption{IID setting}
  \end{subfigure}%
  \begin{subfigure}{.5\columnwidth}
    \centering
    \includegraphics[width=\linewidth]{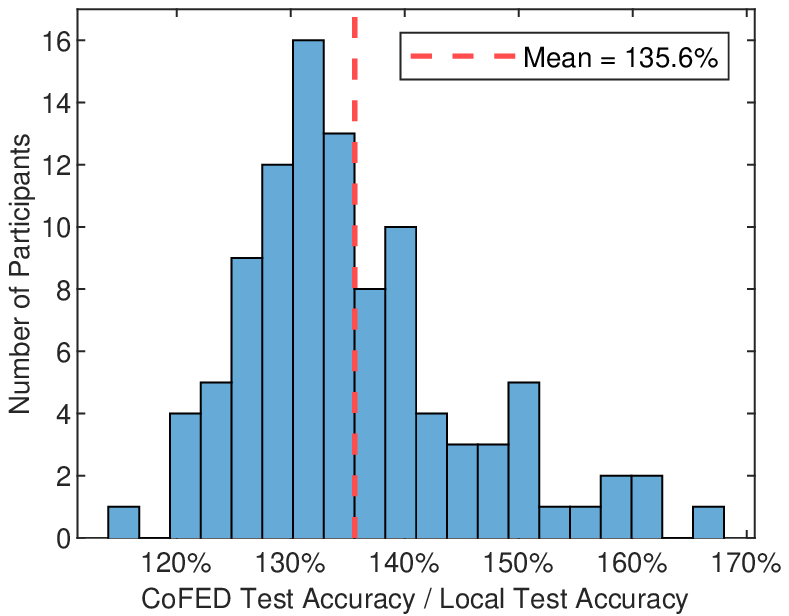}
    \caption{Non-IID setting}
  \end{subfigure}%
  \caption{Results of the CIFAR experiment. The X-axis value is the relative test accuracy, which is the ratio of the CoFED test accuracy to the local test accuracy.}
  \label{cifar_cifar10}
\end{figure}

We test our proposed CoFED method separately under the IID setting and non-IID setting, and the hyperparameter $\alpha$ is set to 0.3 for both settings. We compare the test classification accuracies of the models trained by CoFED with those of the models utilizing local training, and the results are shown in Fig. \ref{cifar_cifar10}(a) and Fig. \ref{cifar_cifar10}(b). Under the IID setting, CoFED method improves the relative test accuracy of each participant model by 10\%-32\% with an average of 17.3\%. This demonstrates that CoFED can increase participant model performance even when the divergences of models are not large (such as those under the IID data setting). For the non-IID setting, CoFED can lead to greater model performance gains due to the greater divergence of data distributions that make locally trained models more divergent. In this experiment, CoFED achieves a relative average test accuracy improvement of 35.6\%, and for each participant, the improvement ranges from 14\% to 67\%. The performance boost under the Non-IID setting is better than the IID setting, which indicates that CoFED suffers less from statistical heterogeneity.

\subsection{FEMNIST Dataset Experiment}
FEMNIST dataset is a handwritten character dataset of LEAF \cite{caldas2018leaf} which is a benchmark framework for FL. It consists of samples of handwritten characters from different users, and we select the 100 users with the most samples of FEMNIST as participants. Forty percent of selected samples are used as training set, and the rest are test set. 

\begin{figure}[ht]
\centering
\includegraphics[width=0.5\columnwidth]{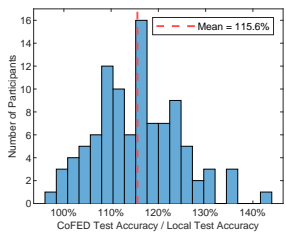}
\caption{Results of the FEMNIST experiment.}
\label{fem_kan}
\end{figure}

The architectures of participant models are the same as those in the CIFAR experiment, and the local training hyperparameters are tuned in similarly to those in the CIFAR experiment. In this experiment, we use random crops of the images in the Chars74k-Kannada dataset \cite{de2009character} to construct an unlabeled public dataset with nearly 50,000 items. Chars74k-Kannada contains handwritten character images of English and Kannada, and only the handwritten Kannada characters are used as unlabeled public dataset in our experiment. The hyperparameter $\alpha$ is set to 0.01, and the results are shown in Fig. \ref{fem_kan}. CoFED improves the relative test accuracies of the models for almost all participants, with an average improvement of 15.6\%.

\subsection{Public Unlabeled Dataset Availability}
One of the major concerns regarding the CoFED method concerns whether a public dataset is available. The role of the public dataset is to express and share the knowledge that the participant models need to learn. Generally, the participants in FL cannot learn knowledge from samples generated entirely by random values, but this does not mean that we must use samples that are highly related to the participants' local data to construct a public dataset.

\begin{figure*}[ht]
\includegraphics[width=\linewidth]{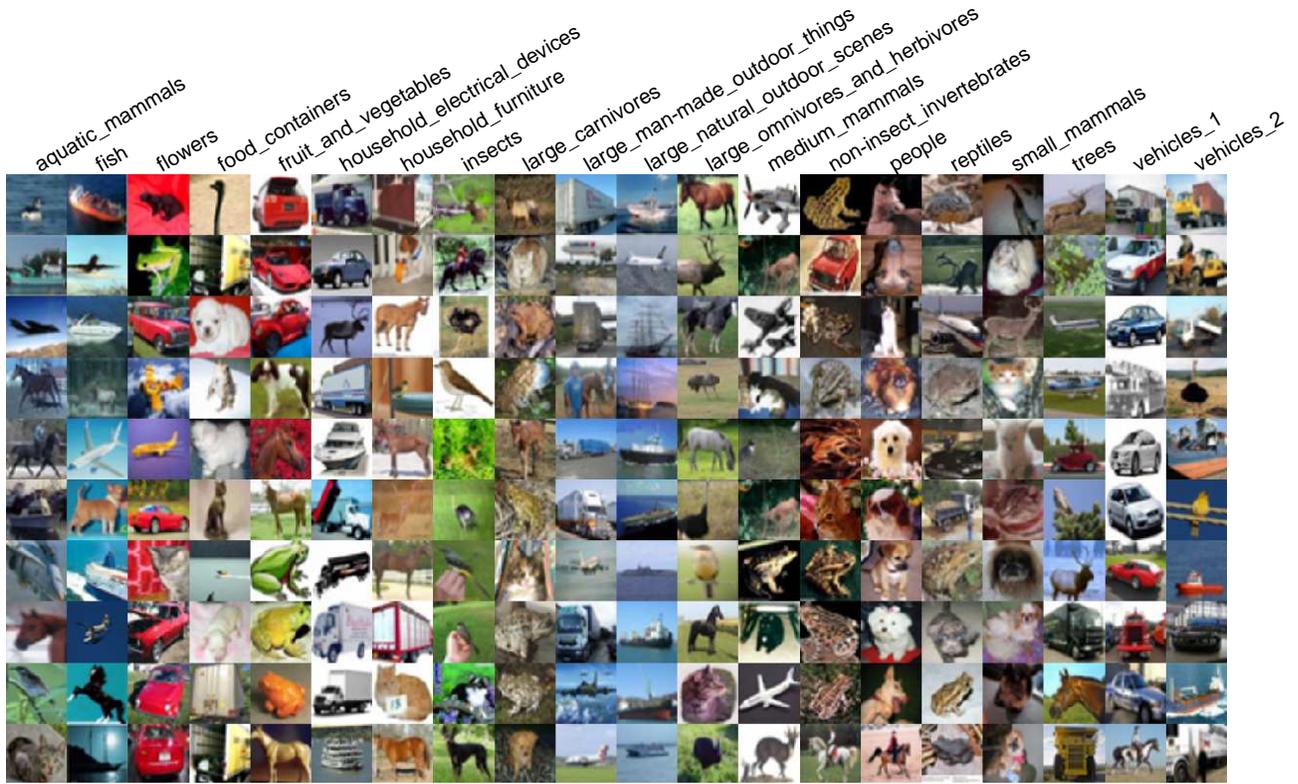}
\caption{The results of pseudolabel aggregation. The images obtained from the training set of CIFAR10 are scattered across the superclasses of CIFAR100. Ten images per superclass are randomly selected.}
\label{pseudo}
\end{figure*}

For example, in the CIFAR experiment, we use the samples of CIFAR100 to construct the local data of the participants, but we use CIFAR10 as the public dataset. The categories of the data samples contained in CIFAR10 and CIFAR100 only overlap slightly. Therefore, we can regard CIFAR10 as a dataset composed of pictures randomly obtained from the Internet without considering the similarity between its contents and the samples of participants (from CIFAR100). A large morphology difference between English characters and Kannada characters is also observed in the FEMNIST experiment. However, CoFED can effectively improve the performance of almost all participant models in both experiments, which makes us want to know how different participant models use public datasets that are not relevant to them to share knowledge. Therefore, we review the results of pseudolabel aggregation and check how these models classify the irrelevant images. Fig. \ref{pseudo} shows a partial example of the pseudolabel aggregation results of the 10 participant models (out of 100 participants).

First, we notice that some trucks and automobiles are correctly classified into the vehicles\_1 category. This indicates that the unlabeled instances whose categories are contained in the label spaces of the participant models are more likely to be assigned to the correct category. However, exceptions occur; some automobiles are assigned to the {\it{flowers}} and {\it{fruit and vegetables}} categories. A common feature possessed by these automobile images is that they contain many red parts, which may be regarded as a distinctive feature of {\it{flowers}} or {\it{fruit and vegetables}} by the classifiers. In addition, the unlabeled samples that are not included in the label spaces of any classifier are also classified into the groups that match their visual characteristics. For example, the corresponding instances of aquatic\_mammals and fish usually have blue backgrounds, which resemble water. Another interesting example is the people category. Although almost no human instances are contained in the training set of CIFAR10, some of the closest instances are still given, including the person on a horse.

\begin{figure}[t]
  \centering
  \begin{subfigure}{.5\columnwidth}
    \centering
    \includegraphics[width=\linewidth]{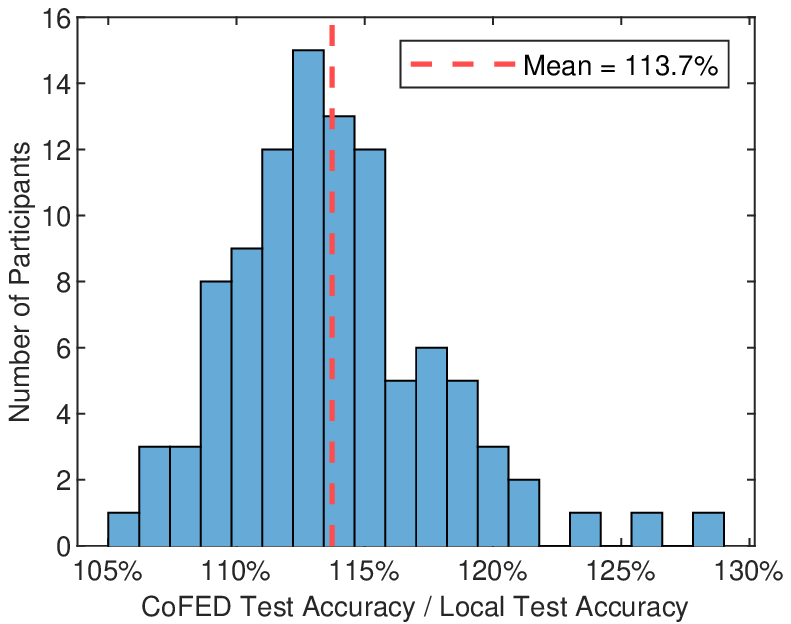}
    \caption{IID setting}
  \end{subfigure}%
  \begin{subfigure}{.5\columnwidth}
    \centering
    \includegraphics[width=\linewidth]{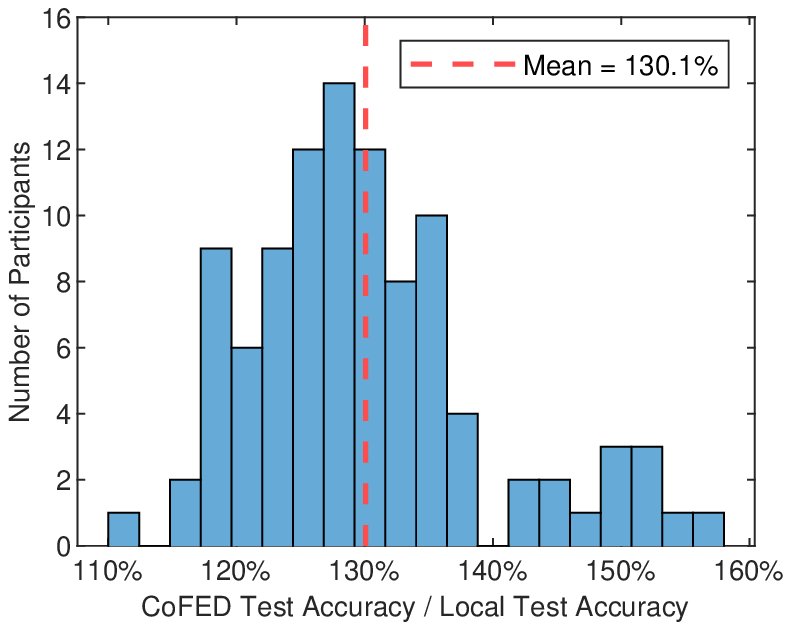}
    \caption{Non-IID setting}
  \end{subfigure}%
  \caption{Results of the CIFAR experiment obtained by using the ImageNet dataset as an unlabeled public dataset.}
  \label{cifar_imagenet}
\end{figure}

Moreover, we also try to replace the public dataset in the CIFAR experiment with the ImageNet dataset \cite{van2016pixel} with almost the same size as that of CIFAR10. The CoFED method can still achieve considerable performance improvements, as shown in Fig. \ref{cifar_imagenet}(a) and Fig. \ref{cifar_imagenet}(b).

\subsection{Unlabeled Dataset Size}
\begin{figure}[ht]
\centering
\includegraphics[width=0.6\columnwidth]{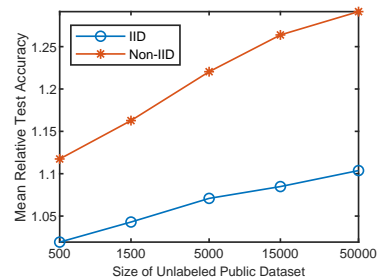}
\caption{Size of the unlabeled public dataset vs. the mean relative test accuracy.}
\label{pubsize}
\end{figure}

According to (\ref{13}), (\ref{14}) and (\ref{15}), when an existing classifier has a higher generalization accuracy and a larger labeled training set, a larger unlabeled dataset is needed to improve its accuracy. This suggests that in the CoFED method, a larger unlabeled dataset can produce a more obvious performance improvement for a given group of participant models. We redo the CIFAR experiment with 10 participants and vary the size of the unlabeled dataset between 500 and 50,000 samples in the training set of CIFAR10. The experimental results are shown in Fig. \ref{pubsize}.

The experimental results are consistent with the theoretical analysis. This shows that the strategy of increasing the size of the unlabeled dataset can be used to boost the performance improvements exhibited by all participant models. Considering that the difficulty of collecting unlabeled data is much lower than that of collecting labeled data in general, this strategy is feasible in many practical application scenarios.

\subsection{Hyperparameter $\alpha$}
From the theoretical analysis, increasing the reliability of the pseudolabeling process is likely to bring more significant performance improvements, which is also very intuitive. Therefore, we use a hyperparameter $\alpha$ to improve the reliability of pseudolabel aggregation. A larger $\alpha$ requires a higher percentage of participants to agree to increase the reliability of the pseudolabels, but this may also reduce the number of available pseudolabel instances. In particular, when the participants disagree greatly, requiring excessive consistency across the results of different participant models may stop the spread of multiparty knowledge.

\begin{figure}[ht]
\centering
\includegraphics[width=0.6\columnwidth]{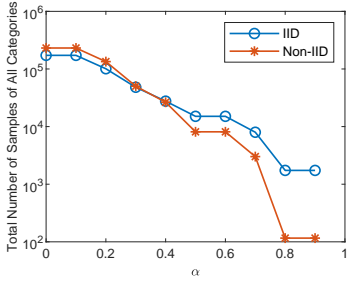}
\caption{$\alpha$ vs. the size of the pseudolabel aggregation results. For $\alpha=1$, the total numbers of IID and non-IID data are both 0, which cannot be shown with the log scale.}
\label{idxSiz}
\end{figure}

\begin{figure}[ht]
\centering
\includegraphics[width=0.6\columnwidth]{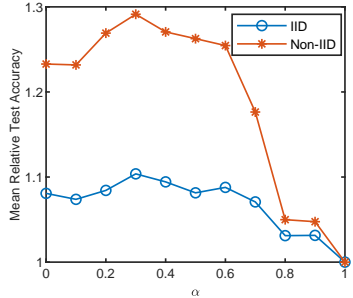}
\caption{$\alpha$ vs. the mean relative test accuracy.}
\label{alphasize}
\end{figure}

In this section, we repeat the CIFAR experiment for 10 participants with different $\alpha$ and record the changes in the total number of samples generated by pseudolabel aggregation when different $\alpha$ values are taken in Fig. \ref{idxSiz}. The changes in the test accuracies of all participant models in the CoFED method are shown in Fig. \ref{alphasize}.

The results show that when the value of $\alpha$ changes, its impact on CoFED is not monotonic. Although an excessively large value of $\alpha$ can increase the reliability of the generated pseudolabels, this also greatly reduces the number of samples in the pseudolabel aggregation results, which may degrade the training results. In the FEMNIST experiment, we find that a larger $\alpha$ value may greatly reduce the number of samples in the pseudolabel aggregation results, so we set the $\alpha$ value to 0.01.

At the same time, an excessively large $\alpha$ value makes the pseudolabel aggregation process more inclined to choose the samples that most participants agree on. Since these sample are approved by most participants, it is almost impossible to bring new knowledge to these participants. In addition, we find that a larger $\alpha$ value has a more severe impact on the non-IID data setting, where performance degradation is more significant than in the IID cases. This is because the differences between the models trained on the non-IID data are greater, and the number of samples that most participants agree on is smaller. Therefore, when $\alpha$ increases, the number of available samples decreases faster than in the IID case, as shown in Fig. \ref{idxSiz}, which causes greater performance degradation under the non-IID setting.

On the other hand, an $\alpha$ that is too small decreases the reliability of the pseudolabel aggregation results, which may introduce more mislabeled samples, making the CoFED method less effective. In addition, an excessively small $\alpha$ value may cause a large increase in the number pseudolabel aggregation samples, resulting in increased computation and communication overheads.

In summary, the effectiveness of the CoFED method can be affected by the value of the hyperparameter $\alpha$, and adopting an appropriate $\alpha$ value can yield greater performance improvements and avoid unnecessary computation and communication costs.

\subsection{Adult Dataset Experiment}
Adult dataset\cite{kohavi1996scaling} is a census income dataset to be used to classify a person's yearly salary based on their demographic data. We set the number of participants to 100 in our Adult experiments. For each participant, we randomly select 200 samples from the training set of the Adult dataset without replacement. Since we try to study the effect of the proposed method on heterogeneous models, 4 types of classifiers are chosen: decision trees, SVMs, generalized additive models, and shallow neural networks. The number of utilized classifiers of each type is 25. The size of unlabeled public dataset in this experiment is 5000, and each sample is randomly generated with valid values for the input properties used by the classifiers. The Adult test set is used to measure the performance of each participant's classifier.

\begin{figure}[ht]
\centering
\includegraphics[width=0.5\columnwidth]{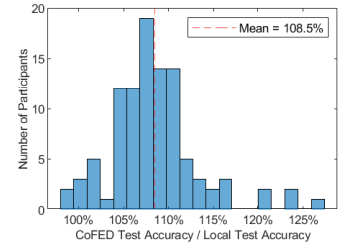}
\caption{Results of the Adult experiment.}
\label{4c}
\end{figure}

The results are shown in Fig. \ref{4c}. The proposed method improves the test accuracies of most classifiers, with an average of 8.5\%, and the performance of the classifier that benefits the most increases by more than 25\%. This demonstrates that the proposed method is applicable not only to image classification tasks involving CNN models but also to nonimage classification tasks with traditional machine learning models.

\subsection{Comparison with Other FL Methods}
To the best of our knowledge, CoFED is the first FL method that tries to be compatible with heterogeneous models, tasks, and training processes simultaneously. Therefore, FL methods are available for comparison under HFMTL settings. We use two well-known comparison methods. One is the personalized FedAvg method, which represents the classic parameter aggregation-based FL strategy that is not compatible with heterogeneous models with different architectures; the other is the FedMD method \cite{li2019fedmd}, which supports different neural network architectures and is used to evaluate the performance of CoFED under heterogeneous models.

To enable the personalized FedAvg and FedMD methods to handle heterogeneous classification tasks, we treat each participant's local label space $\mathcal{Y}_i$ as the union of the spaces $\mathcal{Y}$ in (\ref{union}). In this way, we can handle heterogeneous tasks with the idea of personalized FedAvg and FedMD. The main steps are as follows.
\begin{enumerate}
    \item Use the FedAvg or FedMD algorithm to train a global model whose label space is the union of the label spaces of all participants.
    \item The global model is fine-tuned on each participant's local dataset for a personalized model for their local task..
\end{enumerate}

In this comparison experiment, we use the data configuration presented in Section 4.3. Considering that the personalized FedAvg method can not be used for models with different architectures, 100 participants share the same architecture neural network model. In the FedMD comparison experiment, we use the same setup as that in Section 4.3; that is, we select 100 participants with different neural network architectures.

We try a variety of different hyperparameter settings to achieve better performance in the personalized FedAvg experiment. We use $E=20$ as the number of local training rounds in each communication iteration, $B=50$ is set as the local minibatch size used for the local updates, and participants also perform local transfer learning to train their personalized models in each communication iteration. In the FedMD experiment, We use the similar settings of \cite{li2019fedmd}, and it should be noted that FedMD uses the labeled data of CIFAR10 as the public dataset, which is different from the unlabeled data used in CoFED.

\begin{table}[]
\caption{Comparison Results}
\label{iidtb}
\centering
\begin{threeparttable}
\begin{tabular}{@{}c@{}c@{}ccc}
\hline
                         &          & \begin{tabular}[c]{@{}c@{}}Personalized\\ FedAvg\end{tabular} & FedMD & CoFED \\ \hline
\multirow{2}{*}{IID}     & Accuracy & \textbf{1.06}                                                          & 0.87  & 1.00\tnote{*}  \\ 
                         & Rounds   & 100                                                           & 8\tnote{\dag}     & \textbf{1}     \\ \hline
\multirow{2}{*}{Non-IID} & Accuracy & 0.94                                                          & 0.94  & \textbf{1.00}\tnote{*}  \\ 
                         & Rounds   & 150                                                           & 17\tnote{\dag}    & \textbf{1}     \\ \hline
\end{tabular}
\begin{tablenotes}
\item[*] Reference value.
\item[\dag] The round with the best performance.
\end{tablenotes}
\end{threeparttable}
\end{table}

\begin{figure}[t]
  \centering
  \begin{subfigure}{.5\columnwidth}
    \centering
    \includegraphics[width=\linewidth]{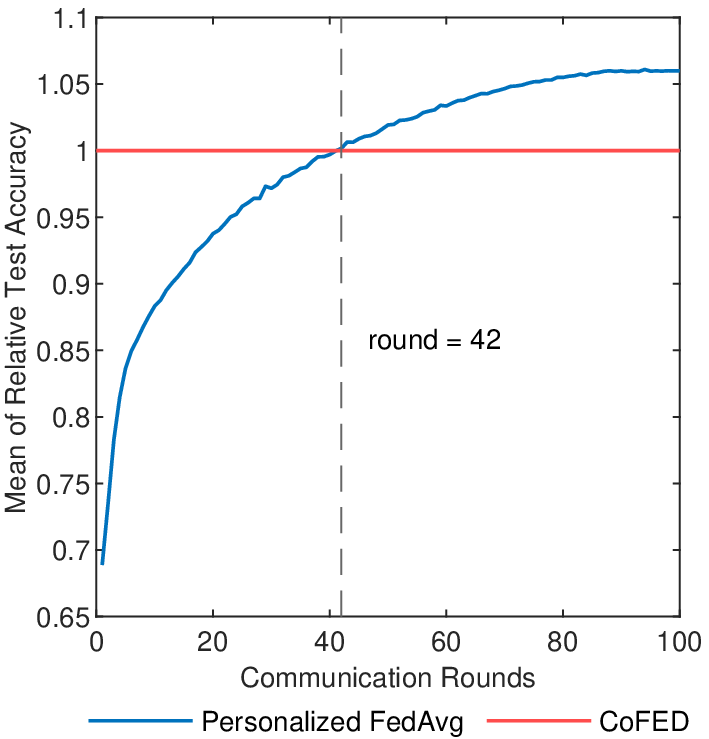}
    \caption{IID setting}
  \end{subfigure}%
  \begin{subfigure}{.5\columnwidth}
    \centering
    \includegraphics[width=\linewidth]{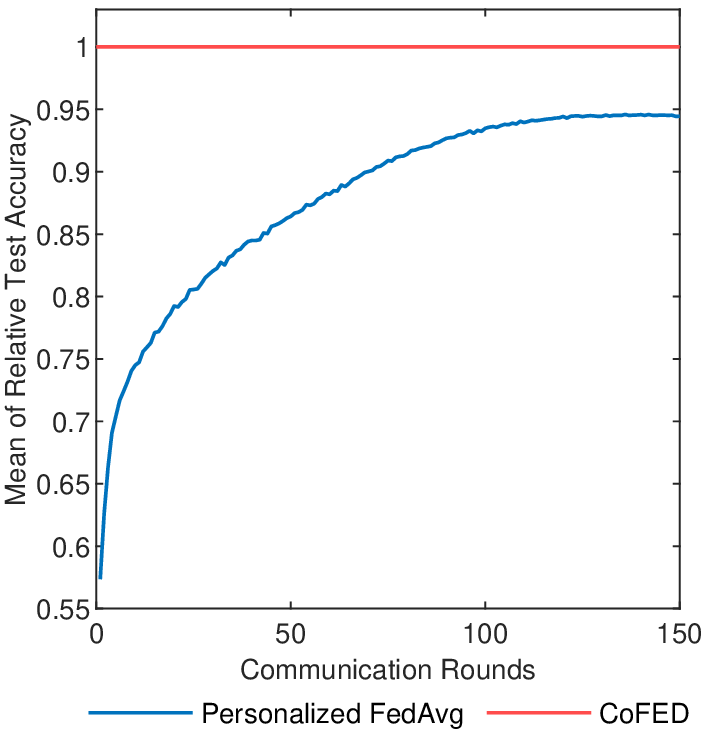}
    \caption{Non-IID setting}
  \end{subfigure}%
  \caption{Personalized FedAvg method vs. CoFED. To facilitate the comparison, the test accuracy of the CoFED model is used as the reference accuracy, and the value of the Y-axis is the ratio of the comparison algorithm's test accuracy to the reference accuracy. Since each participant has a corresponding ratio, the blue line represents the average of the ratios of all participants corresponding to the given number of iterations.}
  \label{fedavg}
\end{figure}

\begin{figure}[t]
  \centering
  \begin{subfigure}{.5\columnwidth}
    \centering
    \includegraphics[width=\linewidth]{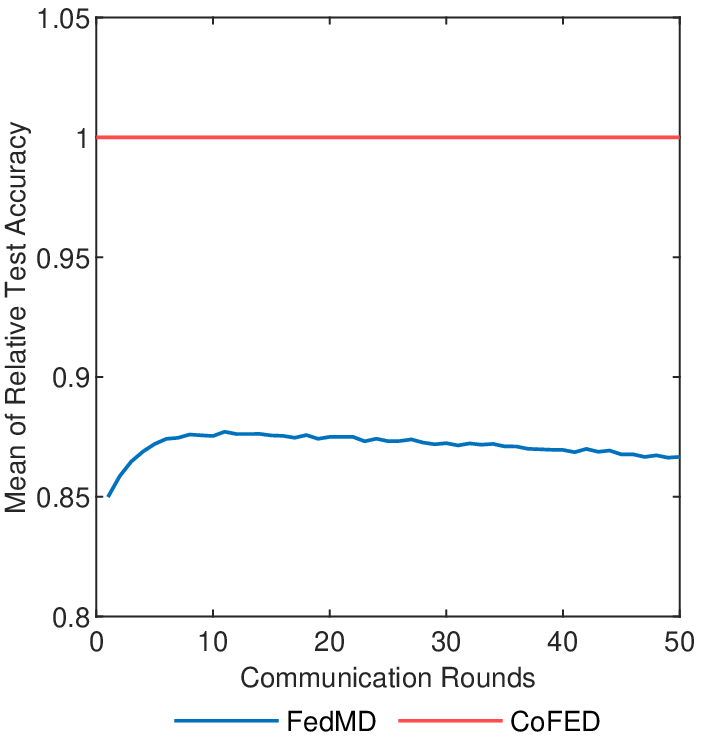}
    \caption{IID setting}
  \end{subfigure}%
  \begin{subfigure}{.5\columnwidth}
    \centering
    \includegraphics[width=\linewidth]{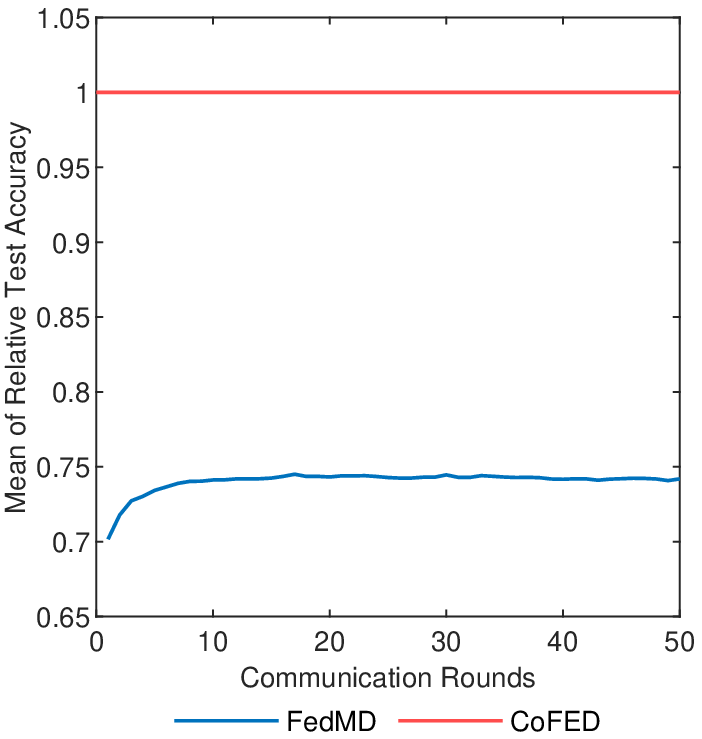}
    \caption{Non-IID setting}
  \end{subfigure}%
  \caption{FedMD vs. CoFED.}
  \label{fedmd}
\end{figure}

The results of the comparison between FedAvg and CoFED are shown in Fig. \ref{fedavg}. The relative test accuracy is calculated as the average of the ratios of all participants to the CoFED test accuracy. Under the IID setting, FedAvg reaches the performance level of CoFED after 42 communication rounds and is finally 6\% ahead of COFED in Fig. \ref{fedavg}(a). Under the non-IID setting, FedAvg stabilizes after 150 communication rounds. At this time, CoFED still leads FedAvg by 6\% in Fig. \ref{fedavg}(b). The comparing results of personalized FedAvg and CoFED are shown in Fig. \ref{fedmd}. For both data settings, CoFED outperforms FedMD, and CoFED leads by 14\% under the IID setting and 35\% under the non-IID one.

In terms of communication cost, if we do not consider the communication overhead required to initially construct the public dataset, CoFED achieves better performance with lower communication overheads in all cases because CoFED only needs to pass through the label data (not the sample itself) during the training process, and iterating for multiple rounds is not required. This assumption is not unrealistic because the construction of public datasets may not require the central server to distribute data to the participants; the participants can instead obtain data from a third party, which does not incur communication costs between the participants and the central server. Even if we include that paradigm, CoFED still achieves better performance with lower communication costs except under the IID and identical architecture model settings. In fact, the IID setting used in the FedAvg comparison experiment is not the scenario that is considered most by CoFED because the model architectures are the same and can be shared under that setting.

\section{Conclusion}
We propose a novel FL method (CoFED) that is simultaneously compatible with heterogeneous tasks, heterogeneous models, and heterogeneous training processes. Compared with the traditional method, CoFED is more suitable for CS-FL settings with fewer participants but higher heterogeneity. CoFED decouples the models and training processes of different participants, thereby enabling each participant to train its independently designed model for its unique task via its optimal training methods. In addition, CoFED protects private data, models and training methods of all participants under FL settings. CoFED enables participants to share multiparty knowledge to increase their local model performance. The method produces promising results under non-IID data settings for models with heterogeneous architectures, which is more practical but is usually difficult to handle in existing FL methods. Moreover, the CoFED method is efficient since training can be performed in only one communication round.

The CoFED method may be limited by the availability of public unlabeled datasets. Although we conduct numerous experiments to demonstrate that CoFED has low requirements for public datasets and that the use of irrelevant or randomly generated datasets is still effective, some failure scenarios may still occur; this is a problem that we hope to address in the future.

\section*{Acknowledgment}
This research was partially supported by the National Key Research and Development Program of China (2019YFB1802800), PCL Future Greater-Bay Area Network Facilities for Large-scale Experiments and Applications (PCL2018KP001).

\bibliographystyle{IEEEtran}
\bibliography{bibs}

\begin{thebibliography}{10}
\providecommand{\url}[1]{#1}
\csname url@samestyle\endcsname
\providecommand{\newblock}{\relax}
\providecommand{\bibinfo}[2]{#2}
\providecommand{\BIBentrySTDinterwordspacing}{\spaceskip=0pt\relax}
\providecommand{\BIBentryALTinterwordstretchfactor}{4}
\providecommand{\BIBentryALTinterwordspacing}{\spaceskip=\fontdimen2\font plus
\BIBentryALTinterwordstretchfactor\fontdimen3\font minus
  \fontdimen4\font\relax}
\providecommand{\BIBforeignlanguage}[2]{{%
\expandafter\ifx\csname l@#1\endcsname\relax
\typeout{** WARNING: IEEEtran.bst: No hyphenation pattern has been}%
\typeout{** loaded for the language `#1'. Using the pattern for}%
\typeout{** the default language instead.}%
\else
\language=\csname l@#1\endcsname
\fi
#2}}
\providecommand{\BIBdecl}{\relax}
\BIBdecl

\bibitem{kairouz2021advances}
P.~Kairouz, H.~B. McMahan, B.~Avent, A.~Bellet, M.~Bennis, A.~N. Bhagoji,
  K.~Bonawitz, Z.~Charles, G.~Cormode, R.~Cummings \emph{et~al.}, ``Advances
  and open problems in federated learning,'' \emph{Foundations and
  Trends{\textregistered} in Machine Learning}, vol.~14, no. 1--2, pp. 1--210,
  2021.

\bibitem{konevcny2015federated}
J.~Kone{\v{c}}n{\`y}, B.~McMahan, and D.~Ramage, ``Federated optimization:
  Distributed optimization beyond the datacenter,'' in \emph{Proceedings of
  workshop on optimization for machine learning in the twenty-ninth conference
  on neural information processing systems (NIPS)}, 2015, pp. 1--5.

\bibitem{rieke2020future}
N.~Rieke, J.~Hancox, W.~Li, F.~Milletari, H.~R. Roth, S.~Albarqouni, S.~Bakas,
  M.~N. Galtier, B.~A. Landman, K.~Maier-Hein \emph{et~al.}, ``The future of
  digital health with federated learning,'' \emph{NPJ digital medicine},
  vol.~3, no.~1, pp. 1--7, 2020.

\bibitem{xiao2021federated}
Z.~Xiao, X.~Xu, H.~Xing, F.~Song, X.~Wang, and B.~Zhao, ``A federated learning
  system with enhanced feature extraction for human activity recognition,''
  \emph{Knowledge-Based Systems}, vol. 229, p. 107338, 2021.

\bibitem{raza2022designing}
A.~Raza, K.~P. Tran, L.~Koehl, and S.~Li, ``Designing ecg monitoring healthcare
  system with federated transfer learning and explainable ai,''
  \emph{Knowledge-Based Systems}, vol. 236, p. 107763, 2022.

\bibitem{courtiol2019deep}
P.~Courtiol, C.~Maussion, M.~Moarii, E.~Pronier, S.~Pilcer, M.~Sefta,
  P.~Manceron, S.~Toldo, M.~Zaslavskiy, N.~Le~Stang \emph{et~al.}, ``Deep
  learning-based classification of mesothelioma improves prediction of patient
  outcome,'' \emph{Nature medicine}, vol.~25, no.~10, pp. 1519--1525, 2019.

\bibitem{adnan2022federated}
M.~Adnan, S.~Kalra, J.~C. Cresswell, G.~W. Taylor, and H.~R. Tizhoosh,
  ``Federated learning and differential privacy for medical image analysis,''
  \emph{Scientific reports}, vol.~12, no.~1, pp. 1--10, 2022.

\bibitem{li2020preserving}
Z.~Li, V.~Sharma, and S.~P. Mohanty, ``Preserving data privacy via federated
  learning: Challenges and solutions,'' \emph{IEEE Consumer Electronics
  Magazine}, vol.~9, no.~3, pp. 8--16, 2020.

\bibitem{gu2021privacy}
B.~Gu, A.~Xu, Z.~Huo, C.~Deng, and H.~Huang, ``Privacy-preserving asynchronous
  vertical federated learning algorithms for multiparty collaborative
  learning,'' \emph{IEEE Transactions on Neural Networks and Learning Systems},
  pp. 1--13, 2021.

\bibitem{zhang2021survey}
C.~Zhang, Y.~Xie, H.~Bai, B.~Yu, W.~Li, and Y.~Gao, ``A survey on federated
  learning,'' \emph{Knowledge-Based Systems}, vol. 216, p. 106775, 2021.

\bibitem{nguyen2021federated}
D.~C. Nguyen, M.~Ding, Q.-V. Pham, P.~N. Pathirana, L.~B. Le, A.~Seneviratne,
  J.~Li, D.~Niyato, and H.~V. Poor, ``Federated learning meets blockchain in
  edge computing: Opportunities and challenges,'' \emph{IEEE Internet of Things
  Journal}, vol.~8, no.~16, pp. 12\,806--12\,825, 2021.

\bibitem{ghimire2022recent}
B.~Ghimire and D.~B. Rawat, ``Recent advances on federated learning for
  cybersecurity and cybersecurity for federated learning for internet of
  things,'' \emph{IEEE Internet of Things Journal}, 2022.

\bibitem{regan2022federated}
C.~Regan, M.~Nasajpour, R.~M. Parizi, S.~Pouriyeh, A.~Dehghantanha, and
  K.-K.~R. Choo, ``Federated iot attack detection using decentralized edge
  data,'' \emph{Machine Learning with Applications}, vol.~8, p. 100263, 2022.

\bibitem{tang2021incentive}
M.~Tang and V.~W. Wong, ``An incentive mechanism for cross-silo federated
  learning: A public goods perspective,'' in \emph{IEEE INFOCOM 2021-IEEE
  Conference on Computer Communications}.\hskip 1em plus 0.5em minus
  0.4em\relax IEEE, 2021, pp. 1--10.

\bibitem{marfoq2020throughput}
O.~Marfoq, C.~Xu, G.~Neglia, and R.~Vidal, ``Throughput-optimal topology design
  for cross-silo federated learning,'' \emph{Advances in Neural Information
  Processing Systems}, vol.~33, pp. 19\,478--19\,487, 2020.

\bibitem{zhang2020batchcrypt}
C.~Zhang, S.~Li, J.~Xia, W.~Wang, F.~Yan, and Y.~Liu, ``Batchcrypt: Efficient
  homomorphic encryption for cross-silo federated learning,'' in \emph{2020
  $\{$USENIX$\}$ Annual Technical Conference ($\{$USENIX$\}$$\{$ATC$\}$ 20)},
  2020, pp. 493--506.

\bibitem{blum1998combining}
A.~Blum and T.~Mitchell, ``Combining labeled and unlabeled data with
  co-training,'' in \emph{Proceedings of the eleventh annual conference on
  Computational learning theory}, 1998, pp. 92--100.

\bibitem{wang2007analyzing}
W.~Wang and Z.-H. Zhou, ``Analyzing co-training style algorithms,'' in
  \emph{European conference on machine learning}.\hskip 1em plus 0.5em minus
  0.4em\relax Springer, 2007, pp. 454--465.

\bibitem{mcmahan2017communication}
B.~McMahan, E.~Moore, D.~Ramage, S.~Hampson, and B.~A. y~Arcas,
  ``Communication-efficient learning of deep networks from decentralized
  data,'' in \emph{Artificial Intelligence and Statistics}, 2017, pp.
  1273--1282.

\bibitem{wang2019adaptive}
S.~Wang, T.~Tuor, T.~Salonidis, K.~K. Leung, C.~Makaya, T.~He, and K.~Chan,
  ``Adaptive federated learning in resource constrained edge computing
  systems,'' \emph{IEEE Journal on Selected Areas in Communications}, vol.~37,
  no.~6, pp. 1205--1221, 2019.

\bibitem{li2020federatedhn}
T.~Li, A.~K. Sahu, M.~Zaheer, M.~Sanjabi, A.~Talwalkar, and V.~Smith,
  ``Federated optimization in heterogeneous networks,'' \emph{Proceedings of
  Machine Learning and Systems}, vol.~2, pp. 429--450, 2020.

\bibitem{smith2017federated}
V.~Smith, C.-K. Chiang, M.~Sanjabi, and A.~S. Talwalkar, ``Federated multi-task
  learning,'' in \emph{Advances in Neural Information Processing Systems},
  2017, pp. 4424--4434.

\bibitem{khodak2019adaptive}
M.~Khodak, M.-F. Balcan, and A.~Talwalkar, ``Adaptive gradient-based
  meta-learning methods,'' in \emph{Proceedings of the 33rd International
  Conference on Neural Information Processing Systems}, 2019, pp. 5917--5928.

\bibitem{lin2020meta}
Y.~Lin, P.~Ren, Z.~Chen, Z.~Ren, D.~Yu, J.~Ma, M.~d. Rijke, and X.~Cheng,
  ``Meta matrix factorization for federated rating predictions,'' in
  \emph{Proceedings of the 43rd International ACM SIGIR Conference on Research
  and Development in Information Retrieval}, 2020, pp. 981--990.

\bibitem{fallah2020personalized}
A.~Fallah, A.~Mokhtari, and A.~Ozdaglar, ``Personalized federated learning with
  theoretical guarantees: A model-agnostic meta-learning approach,''
  \emph{Advances in Neural Information Processing Systems}, vol.~33, pp.
  3557--3568, 2020.

\bibitem{dinh2020personalized}
C.~T. Dinh, N.~H. Tran, and T.~D. Nguyen, ``Personalized federated learning
  with moreau envelopes,'' \emph{Advances in Neural Information Processing
  Systems}, vol.~33, pp. 21\,394--21\,405, 2020.

\bibitem{li2019fedmd}
D.~Li and J.~Wang, ``Fedmd: Heterogenous federated learning via model
  distillation,'' in \emph{Proceedings of NeurIPS 2019 Workshop on Federated
  Learning for Data Privacy and Confidentiality}, 2019, pp. 1--4.

\bibitem{wei2020federated}
K.~Wei, J.~Li, M.~Ding, C.~Ma, H.~H. Yang, F.~Farokhi, S.~Jin, T.~Q. Quek, and
  H.~V. Poor, ``Federated learning with differential privacy: Algorithms and
  performance analysis,'' \emph{IEEE Transactions on Information Forensics and
  Security}, vol.~15, pp. 3454--3469, 2020.

\bibitem{hu2020personalized}
R.~Hu, Y.~Guo, H.~Li, Q.~Pei, and Y.~Gong, ``Personalized federated learning
  with differential privacy,'' \emph{IEEE Internet of Things Journal}, vol.~7,
  no.~10, pp. 9530--9539, 2020.

\bibitem{yin2020fdc}
B.~Yin, H.~Yin, Y.~Wu, and Z.~Jiang, ``Fdc: A secure federated deep learning
  mechanism for data collaborations in the internet of things,'' \emph{IEEE
  Internet of Things Journal}, vol.~7, no.~7, pp. 6348--6359, 2020.

\bibitem{liu2020secure}
Y.~Liu, Y.~Kang, C.~Xing, T.~Chen, and Q.~Yang, ``A secure federated transfer
  learning framework,'' \emph{IEEE Intelligent Systems}, vol.~35, no.~4, pp.
  70--82, 2020.

\bibitem{jia2021blockchain}
B.~Jia, X.~Zhang, J.~Liu, Y.~Zhang, K.~Huang, and Y.~Liang,
  ``Blockchain-enabled federated learning data protection aggregation scheme
  with differential privacy and homomorphic encryption in iiot,'' \emph{IEEE
  Transactions on Industrial Informatics}, vol.~18, no.~6, pp. 4049--4058,
  2021.

\bibitem{fang2021privacy}
H.~Fang and Q.~Qian, ``Privacy preserving machine learning with homomorphic
  encryption and federated learning,'' \emph{Future Internet}, vol.~13, no.~4,
  p.~94, 2021.

\bibitem{li2022blockchain}
D.~Li, D.~Han, T.-H. Weng, Z.~Zheng, H.~Li, H.~Liu, A.~Castiglione, and K.-C.
  Li, ``Blockchain for federated learning toward secure distributed machine
  learning systems: a systemic survey,'' \emph{Soft Computing}, vol.~26, no.~9,
  pp. 4423--4440, 2022.

\bibitem{otoum2022federated}
S.~Otoum, I.~Al~Ridhawi, and H.~Mouftah, ``A federated learning and
  blockchain-enabled sustainable energy-trade at the edge: A framework for
  industry 4.0,'' \emph{IEEE Internet of Things Journal}, 2022.

\bibitem{mo2021ppfl}
F.~Mo, H.~Haddadi, K.~Katevas, E.~Marin, D.~Perino, and N.~Kourtellis, ``Ppfl:
  privacy-preserving federated learning with trusted execution environments,''
  in \emph{Proceedings of the 19th Annual International Conference on Mobile
  Systems, Applications, and Services}, 2021, pp. 94--108.

\bibitem{chen2020training}
Y.~Chen, F.~Luo, T.~Li, T.~Xiang, Z.~Liu, and J.~Li, ``A training-integrity
  privacy-preserving federated learning scheme with trusted execution
  environment,'' \emph{Information Sciences}, vol. 522, pp. 69--79, 2020.

\bibitem{zhao2018federated}
Y.~Zhao, M.~Li, L.~Lai, N.~Suda, D.~Civin, and V.~Chandra, ``Federated learning
  with non-iid data,'' \emph{arXiv preprint arXiv:1806.00582 (2018)}, 2018.

\bibitem{li2020federated}
T.~Li, A.~K. Sahu, A.~Talwalkar, and V.~Smith, ``Federated learning:
  Challenges, methods, and future directions,'' \emph{IEEE Signal Processing
  Magazine}, vol.~37, no.~3, pp. 50--60, 2020.

\bibitem{krizhevsky2009learning}
``A. \text{K}rizhevsky, \text{L}earning multiple layers of features from tiny
  images. \url{https://www.cs.toronto.edu/~kriz/learning-features-2009-TR.pdf},
  2009 (accessed 28 march 2022).''

\bibitem{caldas2018leaf}
S.~Caldas, S.~M.~K. Duddu, P.~Wu, T.~Li, J.~Kone{\v{c}}n{\`y}, H.~B. McMahan,
  V.~Smith, and A.~Talwalkar, ``Leaf: A benchmark for federated settings,''
  \emph{arXiv preprint arXiv:1812.01097 (2018)}, 2018.

\bibitem{de2009character}
T.~E. De~Campos, B.~R. Babu, M.~Varma \emph{et~al.}, ``Character recognition in
  natural images.'' \emph{VISAPP (2)}, vol.~7, p.~2, 2009.

\bibitem{van2016pixel}
A.~Van~Oord, N.~Kalchbrenner, and K.~Kavukcuoglu, ``Pixel recurrent neural
  networks,'' in \emph{International conference on machine learning}.\hskip 1em
  plus 0.5em minus 0.4em\relax PMLR, 2016, pp. 1747--1756.

\bibitem{kohavi1996scaling}
R.~Kohavi \emph{et~al.}, ``Scaling up the accuracy of naive-bayes classifiers:
  A decision-tree hybrid.'' in \emph{Kdd}, vol.~96, 1996, pp. 202--207.

\end{thebibliography}

\end{document}